\documentclass[letterpaper]{article} 
\usepackage{aaai2026}  
\usepackage{times}  
\usepackage{helvet}  
\usepackage{courier}  
\usepackage[hyphens]{url}  
\usepackage{graphicx} 
\usepackage{natbib}  
\usepackage{caption} 
\usepackage{algorithm}
\usepackage{algorithmic}

\usepackage{amsmath}
\usepackage{amssymb}
\usepackage{enumitem}
\usepackage{fancyvrb}

\usepackage{tcolorbox}

\usepackage[dvipsnames,svgnames,x11names]{xcolor}

\usepackage{listings}

\usepackage{newfloat}
\usepackage{listings}
\DeclareCaptionStyle{ruled}{labelfont=normalfont,labelsep=colon,strut=off} 
\floatstyle{ruled}
\newfloat{listing}{tb}{lst}{}
\floatname{listing}{Listing}
\title{Towards Adaptive Environment Generation for Training Embodied Agents}
\author{
    Teresa Yeo\textsuperscript{\rm 1}\equalcontrib,
    Dulaj Weerakoon\textsuperscript{\rm 1}\equalcontrib,
    Dulanga Weerakoon\textsuperscript{\rm 1},
    Archan Misra\textsuperscript{\rm 2}
}

\affiliations{
    \textsuperscript{\rm 1}Singapore-MIT Alliance for Research and Technology Centre\\
    \textsuperscript{\rm 2}Singapore Management University\\ \vspace{0.3em}
    \texttt{\small \{teresa.yeo, dulaj.weerakoon, dulanga.weerakoon\}@smart.mit.edu,
    archanm@smu.edu.sg}
}

\usepackage{bibentry}

\begin{document}

\maketitle

\begin{abstract}
Embodied agents struggle to generalize to new environments, even when those environments share similar underlying structures to their training settings. Most current approaches to generating these training environments follow an open-loop paradigm, without considering the agent's current performance. While procedural generation methods can produce diverse scenes, diversity without feedback from the agent is inefficient. The generated environments may be trivially easy, providing limited learning signal. 
To address this, we present a proof-of-concept for closed-loop environment generation that \textit{adapts difficulty to the agent's current capabilities}. Our system employs a \textit{controllable environment representation}, extracts \textit{fine-grained performance feedback} beyond binary success or failure, and implements a \textit{closed-loop adaptation mechanism} that translates this feedback into environment modifications. This feedback-driven approach generates training environments that more challenging in the ways the agent needs to improve, enabling more efficient learning and better generalization to novel settings.
\end{abstract}


\section{Introduction}

\begin{figure}[t]
    \centering
    \includegraphics[width=0.99\linewidth]{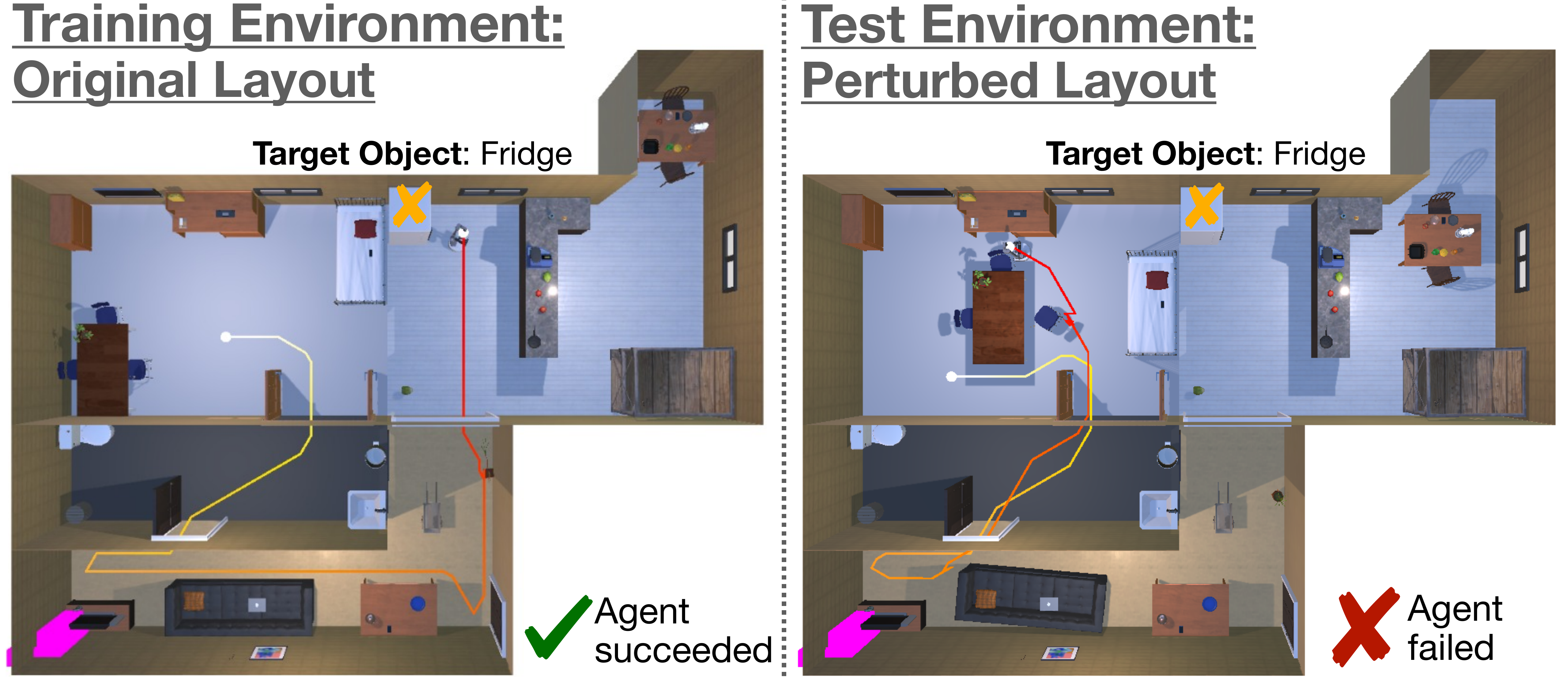}
    \caption{\small \textbf{Embodied navigation performance is sensitive to object perturbations.} Top-down view of agent trajectories (yellow to orange path) for an object navigation task with the fridge as the target object. In the training environment (left), the agent successfully navigates to the target, while in the test environment (right) when the objects and furniture has been rearranged, the agent fails to reach the same target despite starting from the same position. Orange cross indicates target location. Our method analyzes such failure trajectories to generate difficulty-aware environments by programmatically modifying ProcTHOR scene structures, creating meaning and challenging scenarios that can be used to train and improve the agent. Best viewed on screen, zoomed in.}
    \label{fig:problem} \vspace{-1em}
\end{figure}

Embodied agents often struggle to generalize to new environments, even when those environments share similar underlying structures to their training settings. Consider a navigation agent trained in a simple apartment layout. It may successfully navigate from a starting position to a target location, such as a refrigerator, in its training environment. However, when deployed in a visually similar apartment where furniture and objects have been rearranged, the same agent may fail to reach its goal (Fig.~\ref{fig:problem}). Thus, without sufficient diversity in the training data, agents can fail to develop robust navigation strategies that transfer to new settings.

Embodied agents, such as home assistance robots or industrial manipulators, face a data scarcity challenge. 
Unlike disembodied agents like ChatGPT~\cite{chatgpt}, and Gemini~\cite{geminiteam2025geminifamilyhighlycapable} that are trained on massive internet-scale datasets, embodied agents tend to be trained using data generated in simulated physical environments.
Creating these simulations can be expensive and time-consuming, making training data more scarce than for their disembodied counterparts. 
This scarcity represents a bottleneck in developing robust embodied agents.

Most existing approaches to generating training environments follow an open-loop paradigm, i.e., they generate environments \textit{without taking into account the agent's current performance}. Hand-crafted environments can be of high quality but require domain expertise and time to create. Alternatively, procedural generation methods can produce diverse scenes by randomly sampling from templates and object databases. For example, by scanning a real room and generating variations~\cite{deitke2023phone2proc}. While random generation increases diversity, it does not necessarily result in environments that are useful for training e.g., some generated environments may be trivially easy, providing no meaningful learning signal. Thus, \textit{diversity without feedback is inefficient}: we need to generate environments that are appropriately challenging for the given agent we are training.

This paper proposes a proof-of-concept for \textit{closed-loop environment generation} where the difficulty of generated environments is adapted to a given agent's current capabilities. Our system includes three key components. First, we employ a \textit{controllable environment representation} that can be modified efficiently and programmatically. Second, we \textit{extract fine-grained feedback signals} from the agent's performance, going beyond binary success or failure. Third, we implement a \textit{closed-loop adaptation mechanism that translates this feedback into environment modifications}. These components create an adaptive curriculum where training environments grow progressively more challenging in precisely the ways the agent needs to improve, enabling more efficient learning and better generalization to novel settings.

\begin{figure*}[!t]
    \centering
    \includegraphics[width=0.92\linewidth]{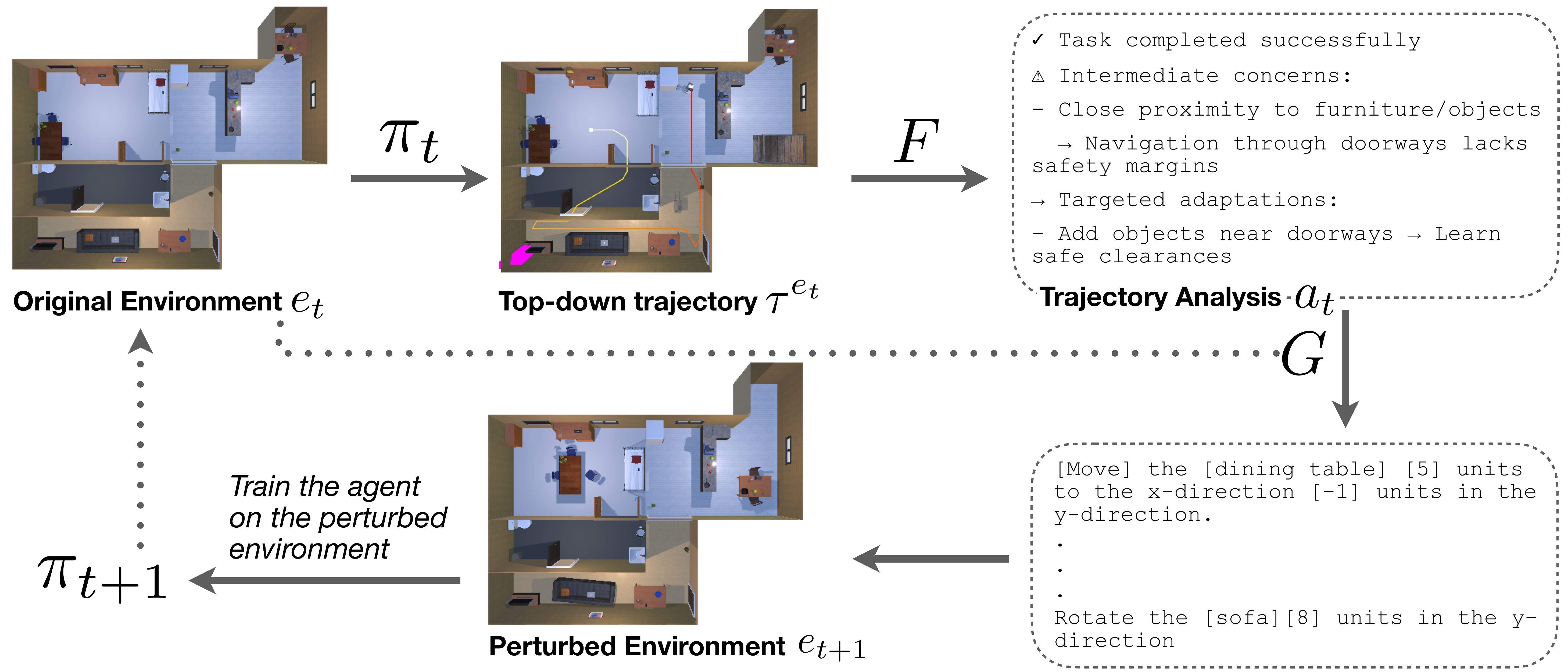}
    \caption{\small \textbf{Overview of the proposed adaptive environment generation framework.} 
    Starting from an original environment $e_t$, an agent with policy $\pi_t$ is deployed to perform embodied tasks (e.g., object navigation), producing a top-down trajectory visualization $\tau^{e_t}$. An analysis model $F$ examines this trajectory to identify success/failure status, intermediate concerns (e.g., unsafe navigation margins), and high-level suggestions for curriculum design. The generator $G$ then translates this analysis $a_t$ into concrete scene graph modifications (e.g., moving or rotating objects) to create a perturbed environment $e_{t+1}$. The agent is retrained on this new environment with policy $\pi_{t+1}$, completing the feedback loop between adaptive environment generation and agent learning.}
    \label{fig:method} \vspace{-1em}
\end{figure*}

\section{Related Work}

\paragraph{Environment Generation Approaches.} 
Environment simulators can be broadly categorized by their generation approach. One category uses scans from real-world environments, e.g., Habitat with the Matterport3D dataset (HM3D)~\cite{ramakrishnan2021habitat}, Replica~\cite{straub2019replica}, ScanNet~\cite{dai2017scannet}, and iGibson~\cite{li2021igibson}, providing photo-realistic reconstructions of actual indoor spaces, from homes to commercial buildings. A second category employs procedural generation techniques to create synthetic environments, as demonstrated by AI2-Thor~\cite{kolve2017ai2}, ProcThor~\cite{procthor}, ThreeDWorld~\cite{gan2020threedworld}, Infinigen~\cite{raistrick2024infinigen}.
Recently, environments can also be generated from natural language descriptions with systems like Genie~\cite{bruce2024geniegenerativeinteractiveenvironments} and Genesis~\cite{Genesis}, allowing for more flexible and user-directed scene creation.

Each approach presents distinct trade-offs. Real-world scans yield highly realistic environments but lack controllability and are less suited for systematic adaptation. Procedurally generated environments offer high controllability at the cost of visual realism. While recent language-based systems like Genie produce realistic scenes, prompting does not allow for the fine-grained control needed for precise scene manipulation. Thus, we selected ProcThor for environment generation due to its balance of controllability and systematic scene composition capabilities.

\paragraph{Open-loop Environment Generation.}
Open-loop generation methods create environments without considering agent performance or task outcomes. These approaches can be further divided into hand-crafted and sampling methods. Hand-crafted approaches like Holodeck~\cite{yang2024holodeck}, LayoutVLM~\cite{sun2025layoutvlm}, and AnyHome~\cite{fu2024anyhome} design individual environments through hand-crafted specifications. Sampling approaches like RoboGen~\cite{wang2023robogen} and Phone2Proc~\cite{deitke2023phone2proc} generate large collections of diverse environments through random sampling procedures. While these open-loop methods can produce diverse training data, they do not leverage feedback from agent's performance to guide environment generation. In contrast, our work employs a closed-loop approach that adapts environment generation based on agent performance, potentially improving data efficiency by focusing on more informative training scenarios.

\paragraph{Closed-loop environment generation}

Closed-loop environment generation automatically creates training environments that adapt to an agent's learning progress.~\citet{kanitscheider2021multi} established learning progress—measured as changes in task success probability—as a reliable signal for curriculum construction in complex domains like Minecraft, demonstrating that dynamically selecting environments based on what the agent can learn next substantially improves training efficiency. Building on adaptive curriculum principles, ADD~\cite{chung2024adversarial} uses diffusion models guided by agent regret to generate diverse adversarial environments that are challenging yet learnable, leveraging the representational power of diffusion models to overcome limitations of parameterized environment spaces. 

Recent work uses foundation models for generating environments: OMNI-EPIC~\cite{faldor2024omni} uses LLMs to generate both environment and reward function code, enabling open-ended creation of any simulatable task while maintaining interestingness and learnability. Similarly, Eurekaverse~\cite{liang2024eurekaverse} applies LLM-based code generation to robotic learning through agent-environment co-evolution, automatically designing terrain curricula for quadrupedal parkour. 
Our work also makes use of these foundation models for closed-loop environment generation.

\section{Method}
We now describe our framework for adaptively generating training environments. See Fig.~\ref{fig:method} for an illustration.

\paragraph{Problem formulation.} Let $e \in \mathcal{E}$ denote an environment configuration, represented by a structured scene graph or parameterized layout. An agent with policy $\pi(a \mid s)$ performs embodied tasks such as object navigation, or manipulation. When deployed on environment $e$, the agent produces a trajectory $\tau^e \in \mathcal{T}$, which can be represented as a sequence of states and actions or as a top-down view (see Fig.~\ref{fig:problem}).

We define a function $F: \mathcal{T} \rightarrow \mathcal{A}$ (e.g., an LLM like GPT-5-mini) that takes the top-down view of trajectory $\tau^e$ and returns an analysis $a \in \mathcal{A}$ describing if the agent succeeded or failed, identifying concerns, and providing high-level suggestions for environment modification.
Next, we define an environment generator $G: \mathcal{E} \times \mathcal{A} \rightarrow \mathcal{E}$ (also an LLM), which at iteration $t$, takes the current environment configuration $e_t$ and the trajectory analysis $a_t = F(\tau^{e_t})$ as input, and outputs a new environment configuration $e_{t+1} = G(e_t, a_t)$.

The goal of $G$ is to produce environments that are both \emph{sufficiently challenging} and \emph{diverse} for the agent. Formally, $G$ aims to \textit{maximize} the following objective: 
$$\mathcal{J}(G) = \mathbb{E}_{t}\left[\Delta R(\pi_t, G(e_t, F(\tau^{e_t})))\right]$$
where $\Delta R$ measures the improvement in agent performance on held-out environments. This formulation establishes a feedback loop between the generator and the agent, allowing for \textit{procedural and adaptive environment generation}.

\paragraph{Structured representation of the environment.} The environment configuration $e$ can be represented as a structured graph capturing the composition of the scene. One such representation is $e = (O, A, R)$, where $O = \{o_i\}$ denotes the set of objects, $A(o_i)$ the attributes of each object (e.g., position, rotation, scale, material), and $R(o_i, o_j)$ the pairwise spatial or functional relations (e.g., \texttt{on}, \texttt{next-to}, \texttt{inside}). This representation can be instantiated in simulators such as AI2-THOR, where an environment is defined by a configuration file specifying object types and their parameters, forming a directed scene graph.

Such a structured representation enables $G$ to perform modifications like adding, removing, or perturbing objects. Furthermore, it also enables a separate validation module to check physical consistency and task feasibility (e.g., ensuring that objects do not overlap and tasks remain solvable). This is discussed at the end of the section.


\paragraph{Fine-grained trajectory analysis as adaptation signal.}
To generate feedback signals for environment adaptation, we employ $F$ to analyze the agent's trajectory i.e., $F$ is given an image like the one on Fig.~\ref{fig:problem}~(left) to analyze. Unlike traditional reinforcement learning approaches that rely on binary task completion signals, our method extracts structured, intermediate-state feedback from the trajectory.
Given the agent's trajectory and task description, $G$ produces:

\begin{itemize}[label={}, leftmargin=0.7em, itemsep=0pt]
\item \textit{Task outcome assessment}: If the agent succeeded or failed.
\item \textit{Intermediate concerns}: Behavioral issues observed during execution (e.g., unsafe clearances, inefficient paths).
\item \textit{Suggestions}: Abstract suggestions that specify what aspects of the environment should change to address the concerns, but not the how (e.g., object placement locations).
\end{itemize}

$F$ maps the trajectory to structured feedback: \{\textit{outcome, concerns, suggestions}\}. The concerns highlight specific states or transitions where the agent's behavior is suboptimal, even when the task succeeds. The adaptations specify abstract environment modifications suggestions to create training scenarios to address these concerns. Note $F$ does not propose low-level suggestions on how to create these scenarios i.e., where to move certain objects.

This approach exploits the rich intermediate information available in embodied navigation trajectories, enabling generation of targeted training environments that addresses specific weaknesses of the agent.

\paragraph{Adaptive environment generation.} 
Given $F$'s analysis $a_t$ of the agent’s trajectory and the environment configuration $e_t$, the generator $G$ aims to produce a more challenging environment configuration $e_{t+1} = G(e_t, a_t)$ for the agent. Concretely, $G$ aims to generate an $e_{t+1} = \arg\min_{e' \in \mathcal{E}} R(\pi_t, e')$ subject to validity and task solvability constraints. 
The agent is then trained in the new environment $e_{t+1}$. This results in an updated policy $\pi_{t+1}$. This interaction between the agent and the generator results in a \textit{curriculum} where the generator continually synthesizes harder environments that are \textit{adapted} to the agent’s current skills.

The generator $G$ can be instantiated in multiple ways depending on the desired level of structure and interpretability. One approach leverages an LLM that conditions on the agent’s trajectory or top-down map of its behavior in $e_t$, and outputs a sequence of discrete editing actions—e.g., selecting an object $o_i \in O$ and modifying its spatial parameters or relations to create a new configuration $e_{t+1}$. As LLMs possess spatial reasoning and world-model priors, this formulation enables broad generalization to unseen scenes and tasks, often producing semantically coherent and meaningful modifications without task-specific training. In contrast, $G$ can also be trained to predict configuration deltas given $e_t$ and the reward $a_t$. This approach supports gradient-based optimization, lower latency, and scalability across many agents or tasks, but may require task-specific data and tends to generalize less reliably beyond the training distribution.


\begin{figure}
    \centering
    \includegraphics[width=0.99\linewidth]{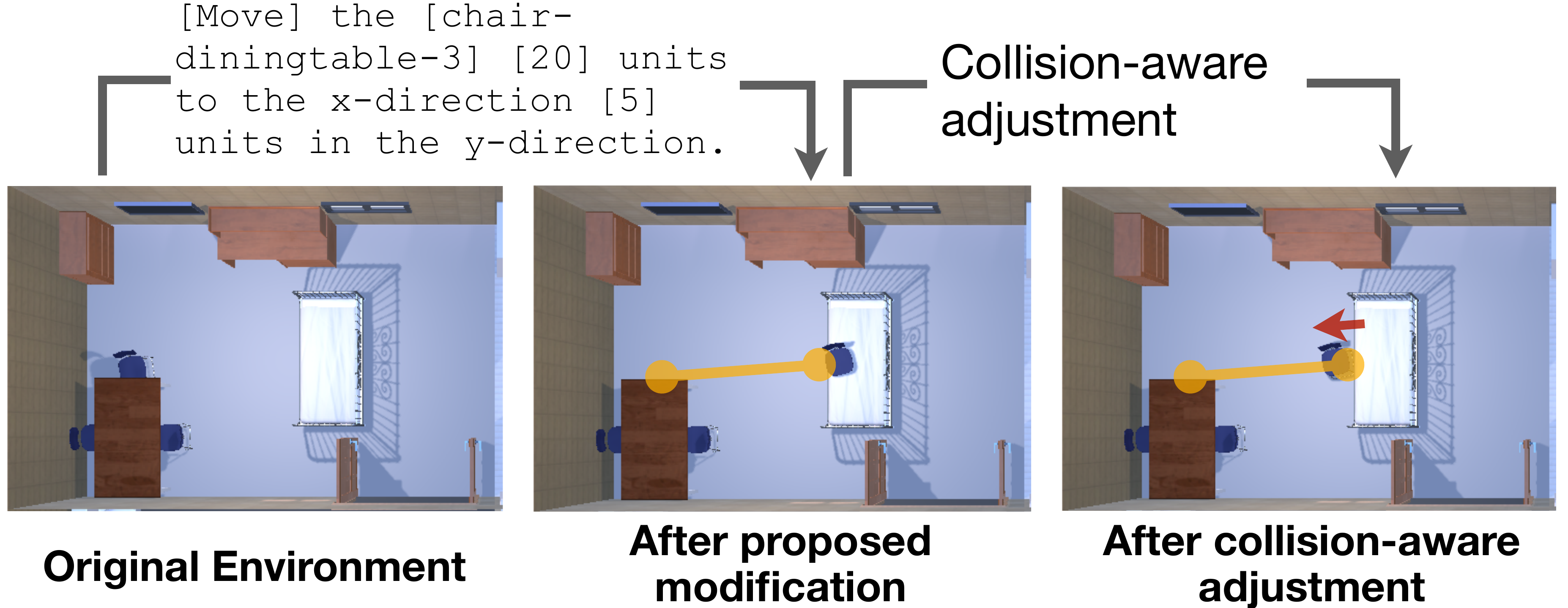}
    \caption{\small \textbf{Collision-aware placements.} When the generator $G$ proposes a new environment configuration that would result in object collisions, we apply a collision-aware adjustment to find a valid placement. In this example, the chair is initially proposed to move 20 units in the x-direction and 5 units in the y-direction, which would cause it to collide with the bed. We compute the displacement vector from the chair's current position to the proposed position and moves the chair incrementally along this direction. At each step, we check for collisions with other objects. This continues until either the proposed position is reached without collision or an obstacle blocks further movement along the path, at which point the chair is placed at the last valid collision-free position.} \vspace{-1em}
    \label{fig:collision-aware-opt}
\end{figure}

\paragraph{Constrained optimization to ensure physical plausibility.} The modifications from $G$ can result in object placements that are not physically plausible e.g., objects intersecting with other objects or walls etc (see Fig.~\ref{fig:collision-aware-opt}). To perform collision-aware placement, we formulate the problem as follows: given an object $o_i^t$ with attributes $\mathcal{A}(o_i^t)$ that include its original position $\mathbf{p}_i^{\text{orig}}$, and a proposed target position $\mathbf{p}_i^{\text{target}}$ generated by $G$, we compute the displacement vector $\mathbf{d} = \mathbf{p}_i^{\text{target}} - \mathbf{p}_i^{\text{orig}}$. We then move $o_i^t$ along this direction in discrete steps of size $\delta$, updating its position component in $\mathcal{A}(o_i^t)$ as $\mathbf{p}_i(k) = \mathbf{p}_i^{\text{orig}} + k\delta\frac{\mathbf{d}}{\|\mathbf{d}\|}$, where $k$ denotes the iteration step, while keeping other attributes (rotation, scale, material) unchanged. 

At each iteration, we perform collision detection between $o_i^t$ and all other objects in the scene. The adjustment process terminates under two conditions: \textbf{1.} $o_i^t$ reaches $\mathbf{p}_i^{\text{target}}$ without collision (i.e., $k\delta \geq \|\mathbf{d}\|$), or \textbf{2.} a collision is detected at step k, in which case we update $\mathcal{A}$ with the last valid collision-free position $\mathbf{p}_i(k-1)$. This approach ensures that the final configuration respects spatial constraints while attempting to satisfy the proposed modification as closely as possible.

\section{Experiments}

\paragraph{Setup.}
We employ GPT-4.1-mini for both the analysis function $F$ and the modification function $G$. For the navigation agent, we utilize a pre-trained policy $\pi$ from Spoc\footnote{\small SigLIP-ViTb-3-CHORESNav-S, downloaded from \url{https://github.com/allenai/spoc-robot-training}}~\cite{ehsani2024spoc}, which has been trained via imitation learning on millions of frames of shortest-path expert trajectories. Given $F$'s trajectory analysis $a$ of policy $\pi$, $G$ modifies the environment one object at a time: it selects an object and proposes a modification (x-y displacement and/or rotation). We use Structured Outputs (via the OpenAI API) to constrain $G$'s response format. Specifically, $G$ selects an object from the list of movable objects in the scene and output: \textbf{1.} the selected object's identifier, \textbf{2.} its x-y displacement values, and \textbf{3.} a rotation angle. This modification is then applied using collision-aware placement. After each modification, the updated top-down environment is rendered and passed to $G$ to generate the next modification. After several steps, we get the final modified environment. See the Appendix for the prompts used with $F$ and $G$.

\paragraph{Preliminary results.} Fig.~\ref{fig:method} shows an example of the perturbed environment $e_{t+1}$. This new environment takes into account the analysis from $F$ by e.g., creating narrower pathways. Furthermore, the perturbed environment is more realistic than randomly perturbing each object. This proof-of-concept demonstrates the feasibility of our approach, thorough comprehensive evaluation including baseline comparisons and agent training on generated environments remains as future work.

\paragraph{Limitations of frontier LLM Spatial Reasoning Capabilities.}
Despite the promise shown in our proof-of-concept, frontier LLMs have known limitations in spatial reasoning. While $G$ can propose modifications in terms of coordinates and rotations, it may struggle to accurately visualize the spatial consequences of these changes, e.g., if moving a sofa 2 units along the x-axis creates its intended effect. The top-down visualization provided to the model helps mitigate this, but LLMs can still propose modifications that are plausible but are spatially incoherent. To address this, we experimented with a verification step: after each modification is rendered, we pass the updated visualization back to $G$ and ask if the change matches its intended effect. If not, $G$ can revise its modification. This mechanism helps catch spatial misunderstandings. 
Note that this differs from the collision-aware placement which ensures physically plausible configurations; the limitation here is with $G$'s mapping of high-level intentions (e.g., creating narrower pathways) to the actual modifications needed to achieve them. These limitations suggest that \textbf{1.} the iterative feedback loop is crucial, allowing $G$ to observe rendered results and correct, and \textbf{2.} future work might benefit from hybrid approaches that combine frontier LLMs with modules that have been trained to perform spatial reasoning for object manipulation.

\section{Conclusion and future work}
This work proposes a closed-loop pipeline for generating new environments for training navigation agents. We assume that the environment can be represented as a structured graph (via simulators like ProcThor). Closed-loop adaptation is then performed by using off-the-shelf LLMs to analyze the agent's trajectory on existing environments and proposing modifications to make this environment more challenging for a given navigation agent. We demonstrate the feasibility of this approach through proof-of-concept implementations. While our results are preliminary, it shows the potential of using LLM-driven closed-loop generation for adaptive environment design. Future work will include comprehensive evaluation against baselines and training navigation agents on the generated environments (over multiple iterations of environment generation and training) to validate the effectiveness of our approach.

\clearpage
\section{Acknowledgments}
This research is supported by the National Research Foundation (NRF), Prime Minister’s Office, Singapore under its Campus for Research Excellence and Technological Enterprise (CREATE) programme. The Mens, Manus, and Machina (M3S)
is an interdisciplinary research group (IRG) of the Singapore MIT Alliance for Research and Technology (SMART) centre.

\bibliography{aaai2026}

\clearpage

\appendix
\onecolumn
{\noindent\Large\textbf{Appendix}}\\
Prompts used with the environment generator $G$ and analysis generator $F$ as described in the experimental setup. 

\begin{tcolorbox}[colback=gray!5,colframe=gray!75,title=Prompt used for $F$ to analyze the top-down view of the agent's trajectory,before skip=3pt,after skip=3pt,fontupper=\footnotesize]
\small
\begin{lstlisting}[numbers=none,xleftmargin=0pt,resetmargins=true,basicstyle=\footnotesize\ttfamily]
Analyze the robot navigation paths shown in this floor plan and provide:

- Success assessment: Did the robot complete its navigation task successfully?
- Intermediate concerns: What issues or risks occurred during navigation (e.g., close proximity to obstacles, narrow clearances, suboptimal routing)?
- Targeted adaptations: What specific improvements or training scenarios should be implemented to address these concerns?
\end{lstlisting}
\end{tcolorbox}

The following prompt is used with the current environment's top-down view to guide the environment generator $G$ in moving individual objects.

\begin{tcolorbox}[colback=ForestGreen!5,colframe=ForestGreen!75,title=Prompt used for $G$ to perturb individual objects,before skip=3pt,after skip=3pt,fontupper=\footnotesize]
\begin{lstlisting}[numbers=none,xleftmargin=0pt,resetmargins=true,basicstyle=\footnotesize\ttfamily]
Analysis of agent's trajectory.
[analysis from F]

Given the analysis of the agent's trajectory, make the environment more challenging for object navigation.

Select ONE object visible in the scene and propose a single move (change in x, y coordinates) that:
- Creates navigation challenges based on the analysis suggestions
- Maintains realism (objects in plausible locations)
- Stays within apartment bounds
Use the `propose_move_instruction` function to specify the move.

Constraints:
- Apartment uses normalized 100x100 grid (x:[0,100], y:[0,100])
- Example: sofa at x=27, y=5, rotation=0
- Keep objects inside walls
- Don't block doorways completely (but can reduce clearances)
- Avoid overlapping with other objects
\end{lstlisting}
\end{tcolorbox}

The following function schema is used with the above prompt and passed as a tool to the OpenAI API. Note that movements are specified using directional commands (left/right for x-axis, up/down for y-axis) rather than absolute coordinates.
\begin{tcolorbox}[colback=ForestGreen!5,colframe=ForestGreen!75,title=Function schema used with the above prompt with $G$,before skip=3pt,after skip=0pt,fontupper=\footnotesize]
\begin{lstlisting}[numbers=none,xleftmargin=0pt,resetmargins=true,basicstyle=\footnotesize\ttfamily]
perturb_instruction_tools = [{
    "type": "function",
    "name": "propose_move_instruction",
    "description": (
        "Propose the change in position for exactly one object. "
        "Movement must be expressed in a normalized 100x100 apartment grid as left/right and up/down units."),
    "parameters": {
        "type": "object",
        "properties": {
            "object_id": {"type": "string", "enum": allowed_object_ids},
            "x_direction": {"type": "string", "enum": ["left", "right"]},
            "x_units": {"type": "number", "minimum": 0, "maximum": 100},
            "y_direction": {"type": "string", "enum": ["up", "down"]},
            "y_units": {"type": "number", "minimum": 0, "maximum": 100},
            "rotation": {"type": "number", "minimum": 0, "maximum": 360}},
        "required": ["object_id", "x_direction", "x_units", "y_direction", "y_units", "rotation"],
        "additionalProperties": False
    }}]
\end{lstlisting}
\end{tcolorbox}

\end{document}